\documentclass[11pt,a4paper]{article}
\usepackage{times,latexsym}
\usepackage{url}
\usepackage[T1]{fontenc}

\usepackage[acceptedWithA]{tacl2021v1}

\usepackage{xspace,mfirstuc,tabulary}

\newif\iftaclinstructions
\taclinstructionsfalse 
\iftaclinstructions
\renewcommand{\confidential}{}
\renewcommand{\anonsubtext}{(No author info supplied here, for consistency with
TACL-submission anonymization requirements)}
\newcommand{\instr}
\fi

\iftaclpubformat 

\else

\fi

\usepackage{times}
\usepackage{latexsym}
\usepackage{import}

\usepackage[T1]{fontenc}

\usepackage[utf8]{inputenc}

\usepackage{microtype}
\usepackage{graphicx}
\usepackage{tabularx}
\usepackage{booktabs}
\usepackage{multirow}
\usepackage{soul}

\usepackage{amsmath}
\usepackage{cleveref}
\usepackage[linesnumbered,ruled,vlined]{algorithm2e}
\definecolor{darkpastelred}{rgb}{0.76, 0.23, 0.13}
\usepackage{amsfonts}
\usepackage{xcolor,colortbl}

\usepackage{subcaption}

\usepackage{tabu}

\usepackage{epstopdf}

\title{Meta-Learning the Difference: Preparing\\ Large Language Models for Efficient Adaptation}

\author{Zejiang Hou \\
  Princeton University\thanks{\; Work done during an internship at Amazon AWS AI.} \\
  \texttt{\small zejiangh@princeton.edu} \\\And
  Julian Salazar \\
  Amazon AWS AI \\
  \texttt{\small julsal@amazon.com} \\\And
  George Polovets \\
  Amazon AWS AI \\
  \texttt{\small polovg@amazon.com}}

\date{}

\begin{document}
\maketitle
\begin{abstract}
Large pretrained language models (PLMs) are often domain- or task-adapted via finetuning or prompting. Finetuning requires modifying all of the parameters and having enough data to avoid overfitting while prompting requires no training and few examples but limits performance. Instead, we prepare PLMs for data- and parameter-efficient  adaptation by  \textit{learning to learn the difference} between general and adapted PLMs. This difference is expressed in terms of model weights and sublayer structure through our proposed dynamic low-rank reparameterization and learned architecture controller. Experiments on few-shot dialogue completion, low-resource abstractive summarization, and multi-domain language modeling show improvements in adaptation time and performance over direct finetuning or preparation via domain-adaptive pretraining. Ablations show our task-adaptive reparameterization (TARP) and model search (TAMS) components individually improve on other parameter-efficient transfer like adapters and structure-learning methods like learned sparsification.
\end{abstract}

\section{Introduction}
\label{sec:introduction}
Finetuning large pretrained language models (PLMs) on task-specific supervised data has become the default strategy to produce performant models for various NLP tasks (\citealp{dai15semi, howard-ruder-2018-universal, radford19gpt2}, \textit{inter alia}), provided a task has enough training data to be adapted to without overfitting. For few-shot tasks, very large PLMs like the 175B-parameter GPT-3 \citep{brown20gpt3} do surprisingly well without training using \textit{prompts}, where task-specific examples $(x_j,y_j)$ are presented as text to condition the PLM before a test input $x_{\text{test}}$ is given. Our work considers an important middle ground: minimizing the computational cost of finetuning while improving on its performance in low-resource and few-shot settings.

In general, self-supervised objectives used for PLMs assume little about the nature of downstream tasks.
Earlier works suggested that task-awareness is unnecessary for PLMs of sufficient scale; e.g., \citet{raffel20t5} found that multi-task learning underperformed pretrain-finetune for the largest T5 models on multi-format question answering. However, \citet{gururangan-etal-2020-dont} showed that further pretraining on unlabeled text from the downstream task (task-adaptive pretraining, or TAPT) or a related domain (DAPT) consistently improved adaptation performance.  \citet{aghajanyan-etal-2021-muppet} revisited \citeauthor{raffel20t5}\ and found that by greatly improving the number and balance of tasks, one can utilize a multitask objective after pretraining and achieve gains in proportion to the number of tasks. As for even larger models, \citet{brown20gpt3} argue that the impressive few-shot prompting ability of GPT-3 comes from ``implicit'' meta-learning \cite{schmidhuber87evolutionary, bengio90learning} which they term \textit{in-context learning}, where the outer loop is performed by self-supervised pretraining, and the inner loop is performed by forward passes on implicit examples in unlabeled texts.

\begin{figure}[t]
\small
\captionsetup{font=small}
\captionsetup[sub]{font=small}
\centering
\begin{subfigure}[b]{\linewidth}
  \includegraphics[width=1\columnwidth,trim={0 1cm 0 0},clip]{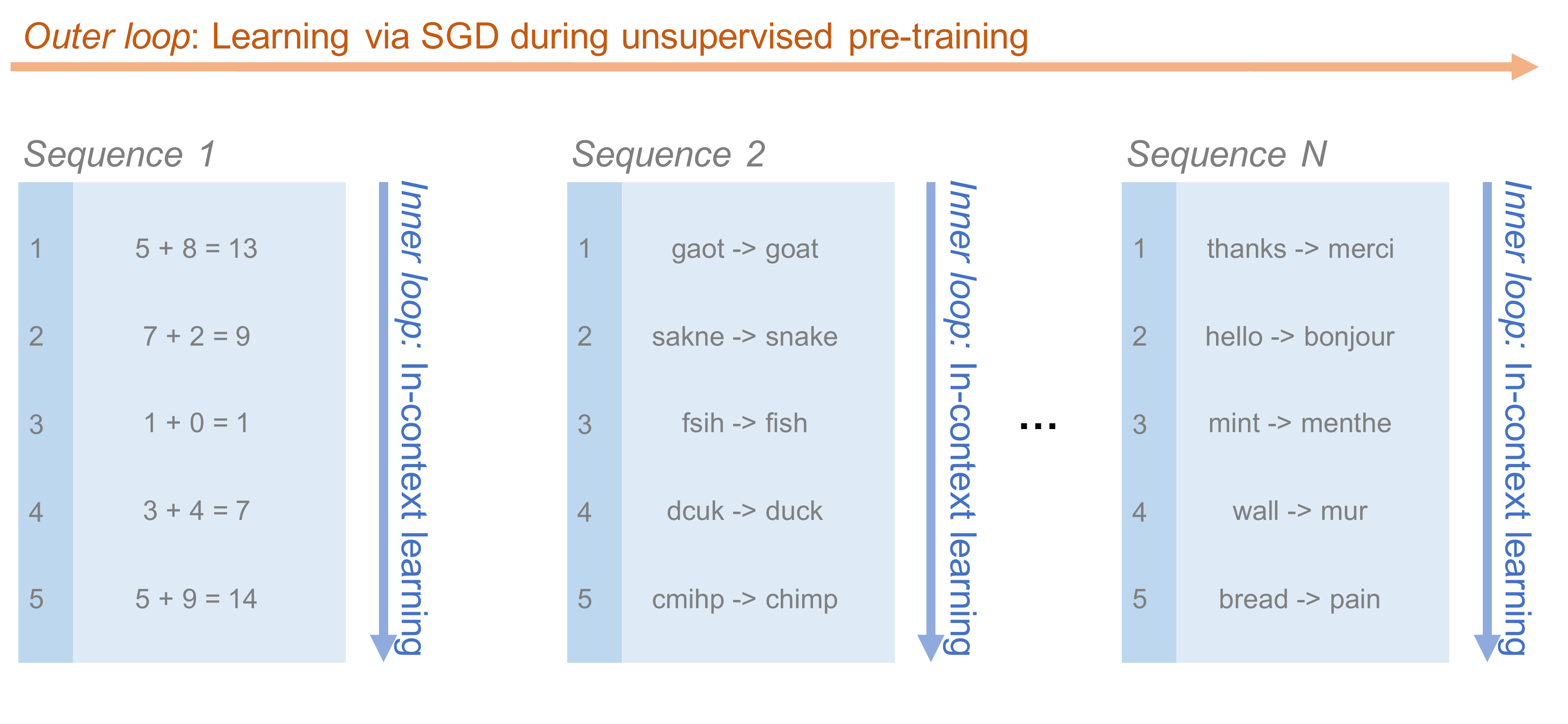}
\end{subfigure}
\par\medskip
\begin{subfigure}[b]{\linewidth}
\centering
  \includegraphics[width=1\columnwidth]{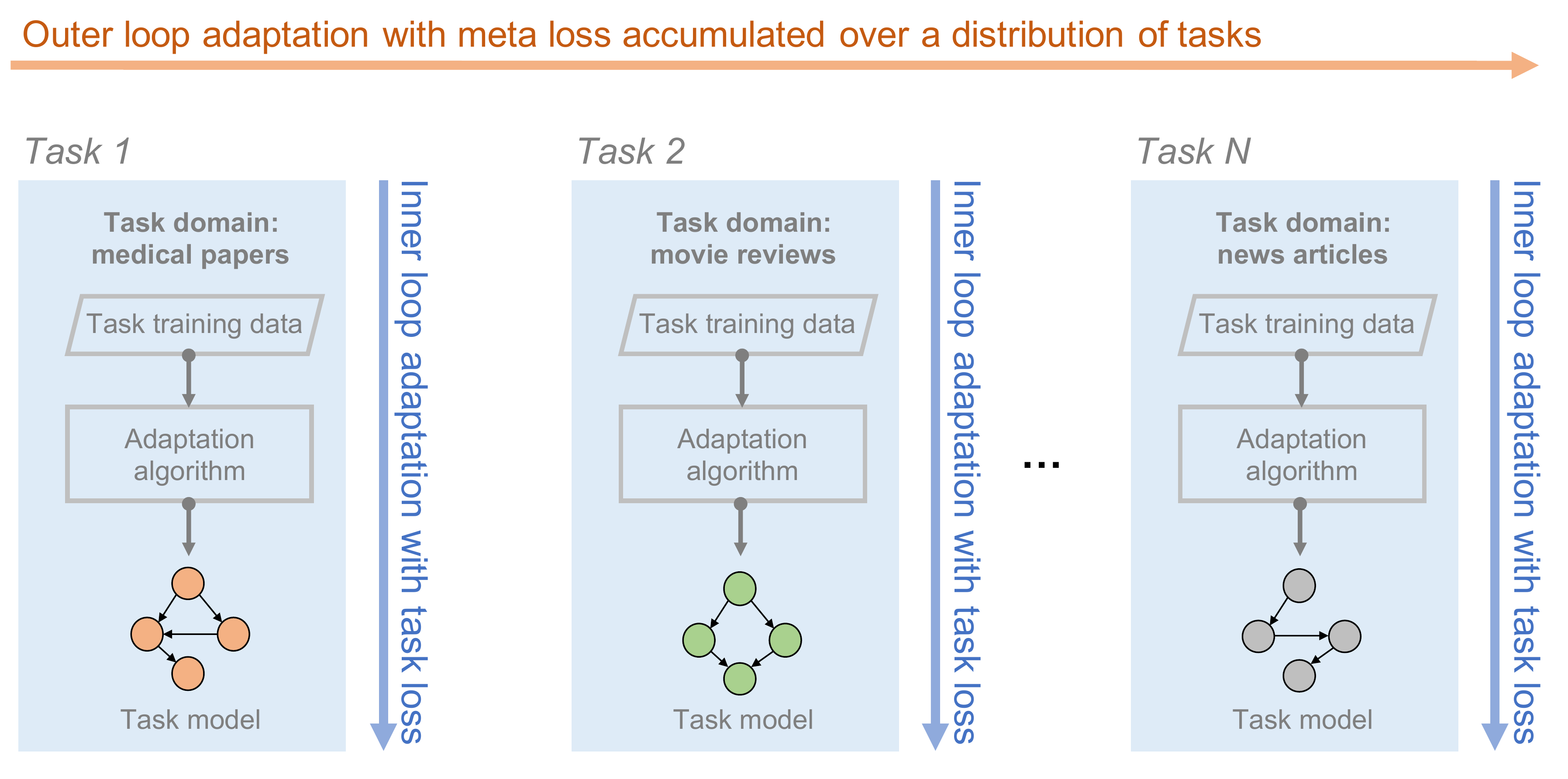}
\end{subfigure}
\caption{Comparison between (\textbf{top}) \textit{implicit} meta-learning from text corpora that incidentally contain task ``prefixes'', as in GPT-3 (\citealp{brown20gpt3}; Fig.\ 1.1), and (\textbf{bottom}) \textit{explicit} meta-learning the transformation of a PLM's weights and sublayers for a distribution of tasks.}
\label{fig:scheme}
\end{figure}

These works motivate that exposure to broad information about downstream tasks remains useful in preparing a large PLM for adaptation. Hence, {we propose explicit meta-learning for preparing large PLMs for data-efficient adaptation}; a visual comparison is in \Cref{fig:scheme}. To also achieve parameter efficiency and performance, {we adapt meta-transfer learning} \citep{sun19meta} {to large PLMs in two proposed ways}: an inner loop optimizing a low-rank {task-adaptive reparameterization (TARP)} of weights, and an outer loop learning an architecture controller for searching {task-adaptive model structures (TAMS)}. These improve over general finetuning and even DAPT-prepared LMs on generative and unconditional few-shot and low-resource settings, such as multi-domain abstractive summarization (AdaptSum; \citealp{yu-etal-2021-adaptsum}) and language modeling.

Furthermore, our analysis shows that {each component of our task distribution-aware strategy independently improves over prior work}: (1) meta-transfer learning improves over model-agnostic meta learning \citep{finn17maml} even after multitask learning on the same data, setting a new state-of-the-art on few-shot Persona-Chat dialog personalization \citep{zhang-etal-2018-personalizing}; (2) our proposed dynamic low-rank TARP outperforms recent methods such as MAM adapters \citep{he2022towards} and alternate reparameterizations like Kronecker products \citep{zhang21beyond}; (3) our lightweight controller for generating task-aware architectures in TAMS extends improvements into higher resource tasks and rediscovers task-specific modifications like 1D convolutions for Transformers.

\begin{figure*}[ht!]
\small
\captionsetup{font=small}
\captionsetup[sub]{font=small}
\centering
\includegraphics[width=\linewidth]{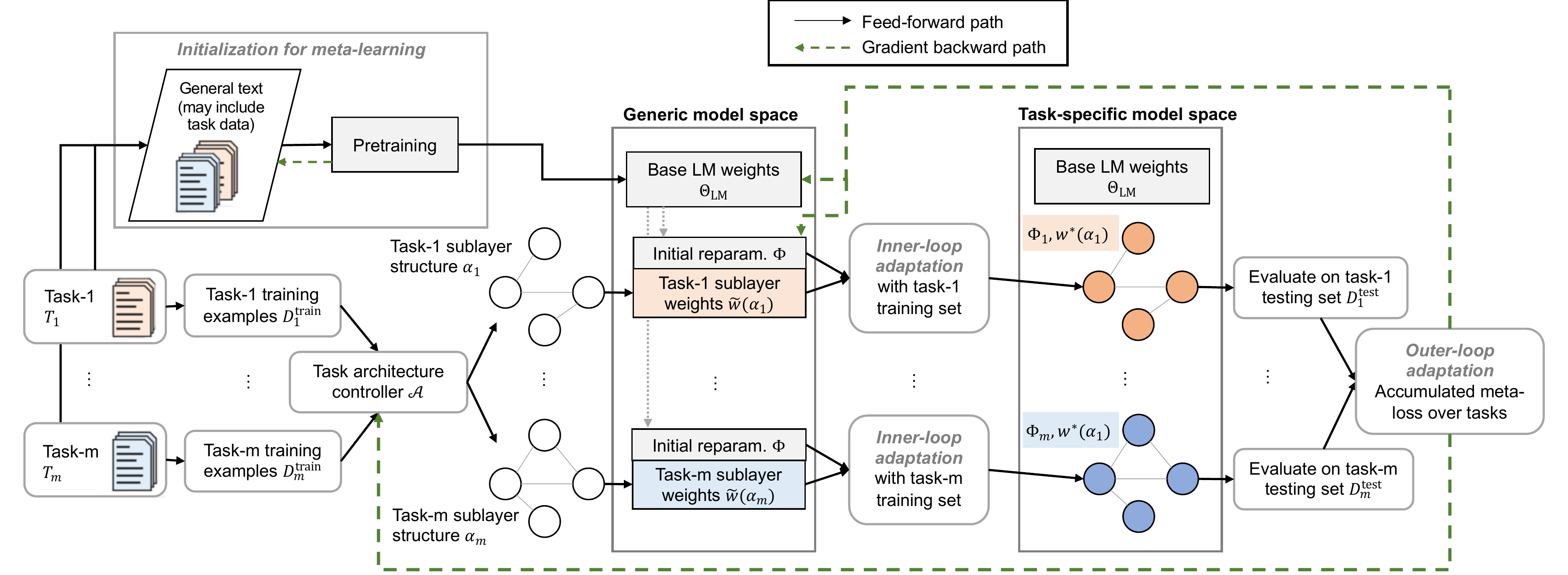}
\caption{Overview of our proposed method, which learns to transform a small set of weights $\Phi_i$ (TARP learning) and modify sublayer modules $\alpha_i$ (TAMS learning) in a task-specific, data-efficient, and parameter-efficient manner.
First, we initialize with a base PLM (\textbf{top left}).
In each meta-iteration, we sample a batch of tasks from a task distribution (\textbf{left}).
In the inner loop (\textbf{middle}), independent sets of dynamic low-rank reparameterizations are initialized, and an architecture controller generates independent task-specific sublayer modules, all of whose weights are adapted to the task's training set.
Each task model is evaluated on the corresponding task's test set. 
In the outer loop (\textbf{right}), these task losses are summed up to produce the overall meta-loss, and the backward path optimizes the base model, the initial reparameterization, and the architecture controller.
}
\label{fig:mltd}
\end{figure*}

Our proposal is summarized in \Cref{fig:mltd}, with pseudocode in Algorithm \ref{algorithm: overview} at the end of the next section. We publicly release the code for our experiments and our reference library online.\footnote{\url{https://github.com/amazon-research/meta-learning-the-difference}}

\section{Methodology}
\label{sec:meta-learning}

Our goal is to explicitly optimize a PLM for efficient adaptation to any task $\mathcal{T}_i$ sampled from a distribution of low-resource NLP tasks $p(\mathcal{T})$. Each task consists of a training set $\mathcal{D}_{i}^{\text{train}}$, a test set $\mathcal{D}_{i}^{\text{test}}$, and a loss function $\mathcal{L}_i$.

The prevailing approach for efficiently optimizing a base model $f_{\Theta}$ on a (relatively) small task-specific dataset is to use model-agnostic meta-learning (MAML;  \citealp{finn17maml}). This is a bi-level optimization process that uses a stochastic gradient-based strategy to sample a batch of tasks $\{\mathcal{T}_i\}_{i=1}^{B}$ from the task distribution $p(\mathcal{T})$  in each meta-iteration. In the inner loop, each task finetunes a copy of the model's weights $\Theta$ for a small number of steps $T_\text{in}$, producing task-specific weights $\Theta_i$. In the outer loop, each task model $f_{\Theta_i}$ is evaluated on its corresponding task's test set $\mathcal{D}_{i}^{\text{test}}$ and these losses are summed to produce the overall meta-loss. The meta-loss  $\sum_{\mathcal{T}_i \sim p(\mathcal{T})}\mathcal{L}_{i}^{\text{test}}(f_{\Theta_i})$ is then used to optimize and update $\Theta$; see \citet{metalearningblog}\footnote{\url{https://lilianweng.github.io/lil-log/2018/11/30/meta-learning.html}} for a more detailed overview.

MAML, however, is not generally used in NLP as a competitive alternative to pretrain-then-finetune methods for low-resource and few-shot settings.
To rectify MAML's limitations, we propose a meta-learning the difference (MLtD) framework to optimize PLMs for fast and data-efficient adaptations with the following contributions:

\paragraph{MAML after pretraining.} Earlier works performed MAML using random initializations, or at best with pretrained token embeddings \citep{madotto-etal-2019-personalizing}, which was shown to underperform the pretrain-finetune paradigm. With the increased prevalence of large-scale pretraining, 
recent works have begun to initialize MAML with PLMs \citep{dou-etal-2019-investigating}. 
We continue this approach, but further show that pretraining + MAML, even when labeled (i.e., multitask) and performed only on the meta-training data (i.e., no external text), improves performance and mitigates overfitting versus pretraining alone or MAML alone (\Cref{sec:analysis}), suggesting that pretraining produces a better initialization that promotes generalization in later meta-learning.

\paragraph{Parameter-efficient transfer.} Adaptation data is typically limited, making it easy for large models to overfit. 
Previous works use very shallow CNNs \citep{finn17maml}, only adapt scale-and-shift parameters atop the original model \citep{sun19meta}, or apply various general regularization techniques such as weight decay, label smoothing, dropout, early stopping, and $\ell_1$ regularization \citep{madotto-etal-2019-personalizing,song-etal-2020-learning}.
In contrast, we propose learning 
dynamic low-rank reparameterizations $g_{\Phi_i}$ (\Cref{sec:parameter-efficient}) of the base model such that $\Theta_i(x) = g_{\Phi_i}(\Theta_{\text{LM}}, x)$ for task $\mathcal{T}_i$.
Here, $\Phi$ is a small set of new parameters that are adapted into task-specific $\Phi_i$ when finetuning.
Notably, we modify MAML to incorporate these parameter-efficient modules, so that during task adaptation in both meta-training (the inner loop) and meta-testing (novel tasks) $\mathcal{T}_i$, we only adapt $\Phi \to \Phi_i$ instead of $\Theta \to \Theta_i$, speeding up both phases and improving overall performance. 
Though some works explore the benefits of joint training or fusion  of parameter-efficient modules \citep{stickland19pals,lin-etal-2020-exploring, pfeiffer-etal-2021-adapterfusion}, 
prior work has not explored meta-learning to learn these adaptations in a task distribution-aware setting.

\paragraph{Architecture adaptation.}
While the Transformer has proven to be a robust general-purpose architecture, recent work has shown that the optimal attention-then-FFN sublayer structure can vary across tasks (e.g., Sandwich Transformers;  \citealp{press-etal-2020-improving}).
However, previous data-driven sublayer searches are often task-agnostic (e.g., \citealp{so2021primer}), where the sublayer search is implemented before pretraining.
Meta-learning enables learning data-driven sublayers \textit{after} pretraining, in a differentiable, task-adaptive manner (\Cref{sec:task-adaptive}).
Instead of a separate search per task as in previous methods (DARTS; \citealp{liu19darts}), we propose meta-learning a task-aware architecture controller to help it generalize to new tasks when searching neural architectures, by learning to directly generate task-specific sublayer structures from the dataset.
By exploiting architectural knowledge learned over the task distribution, our task-adaptive model structure approach improves test-time performance.
A related work in customizing model structure is CMAML \citep{song-etal-2020-learning}, which
applies a sparse pruning algorithm to obtain task-specific weight masks.
Our method differs in that we consider generalization over a distribution of tasks (instead of a single task), and has a richer search space with different operations, numbers of layers, and widths of layers, so that our method provides architecture diversity to accommodate to the different task data.
\\

In all, we employ meta-learning to improve upon initializing from a pretrained $\Theta_{\text{LM}}$, allowing better downstream finetuning on tasks. 
By learning only the transformation weights $\Phi_i$ and 
(optionally) the task-specific architecture $\alpha_i$ 
for new tasks $\mathcal{T}_i$, our method ``learns to learn the difference'' between a PLM and a task-specific LM in a training-efficient way.

\subsection{Efficient parameter adaptation}
\label{sec:parameter-efficient}
We categorize recent works in parameter-efficient adaptation of large PLMs into three types:

\paragraph{Adding parameter-efficient layers.} Low-dimensional adapters \citep{rebuffi18adapters} have been injected into a frozen pretrained BERT either serially after each sublayer \citep{houlsby19adapters}, or in parallel to the self-attention layers (PALs; \citealp{stickland19pals}). 
Following works \citep{bapna-firat-2019-simple,lin-etal-2020-exploring} applied adapters to other NLP models, e.g. GPT-2.
Compacters \citep{mahabadi21compacter} reduce the adapter parameter count via hypercomplex multiplications \citep{zhang21beyond}. 

\paragraph{Adding parameter-efficient prefixes.} Inspired by prompting, the learning of automated prompts or task-specific continuous variants has been applied for encoder-only PLMs like BERT \citep{shin-etal-2020-autoprompt, hambardzumyan-etal-2021-warp} and generative PLMs \citep{li-liang-2021-prefix, liu21ptuning, lester-etal-2021-power} where one learns task-specific vectors prepended to inputs or hidden representations.

\paragraph{Transformations only.} The adapter and prefix-tuning strategies insert layers or introduce prefixes, increasing inference time or in-memory size. Instead, \citet{zhao-etal-2020-masking} learn binary masks, and diff pruning \citep{guo-etal-2021-parameter} learns sparse additive vectors. Both methods use unstructured sparsity to achieve parameter efficiency.
Later works like BitFit \citep{zaken21bitfit} and LoRA \citep{hu21lora} introduce parameter-efficient modifications targeting the Transformer architecture:
BitFit only tunes the bias parameters, while LoRA adds low-rank decomposition weights
to the self-attention weights.
Recently, \citet{he2022towards} proposed parallel mix-and-match (MAM) adapters which leverage benefits of the preceding types.\\

Hence, to minimize overhead we focus on a ``transformations only'' approach.
Inspired by the scale-and-shift parameters of \citet{sun19meta}, we propose learning \textit{affine} transformations to reparameterize the pretrained model weights towards a task.
For a pretrained weight matrix $W_0^l\in\mathbb{R}^{C_{\text{in}}\times C_{\text{out}}}$ (can be any dense layer in the self-attention module or the FFN module in a transformer based architecture), we first reparameterize the task-specific weights as:
\begin{equation}
    W^l = \Phi_{1}^l\odot W_0^l + \Phi_{2}^l \label{reparameterization}
\end{equation}
where $\Phi_{1}^l, \Phi_{2}^l \in \mathbb{R}^{C_{\text{in}}\times C_{\text{out}}}$ and $\odot$ denotes the elementwise (Hadamard) product.

At adaptation time, we apply low-rank constraints while optimizing the reparameterization weights only, giving the training objective
\begin{equation}
\underset{\substack{\{\Phi_1^l, \Phi_2^l\}_{l=1}^{L}\\ \text{rank}(\Phi_{i}^l)<r}}{\text{min}} \sum_{t=1}^T \log p(y_{t}|x,y_{<t}; {\{W^l\}_{l=1}^L}).
\end{equation}

A straightforward approach to solve the rank-constrained problem is to apply a low-rank decomposition to the transformation weights $\Phi_{i}^{l}$. We term this approach of learning parameter-efficient affine transformations \textbf{task-adaptive reparameterization (TARP)}. We consider two standard static decomposition methods:
\begin{itemize}
    \item \textbf{Bilinear}, which takes $\Phi_{j}^{l} = U_{j}^l{V_{j}^l}^{T}$ where $U_{j}^l\in\mathbb{R}^{C_{\text{in}}\times r}$ and $V_{j}^l\in\mathbb{R}^{C_{\text{out}}\times r}$, as done in the additive-only setting ($\Phi_{1}^l = I$) by LoRA.
    \item \textbf{Kronecker product}, which takes $\Phi_j^l = \sum_{k=1}^n H_k^l\otimes (U^l_k{V^l_k}^{T})$ where $H_k\in\mathbb{R}^{n\times n}$, $U^l_k\in\mathbb{R}^{({C_{\text{in}}}/{n})\times r}$, $V^l_k\in\mathbb{R}^{({C_{\text{out}}}/{n})\times r}$, and $n$ is a hyperparameter, as used in the ``added-layer'' Compacter approach.
\end{itemize}

\begin{figure}[t]
\centering
\small
\captionsetup{font=small}
\captionsetup[sub]{font=small}
\includegraphics[width=0.7\linewidth,trim={0 0.1cm 0 0.5cm},clip]{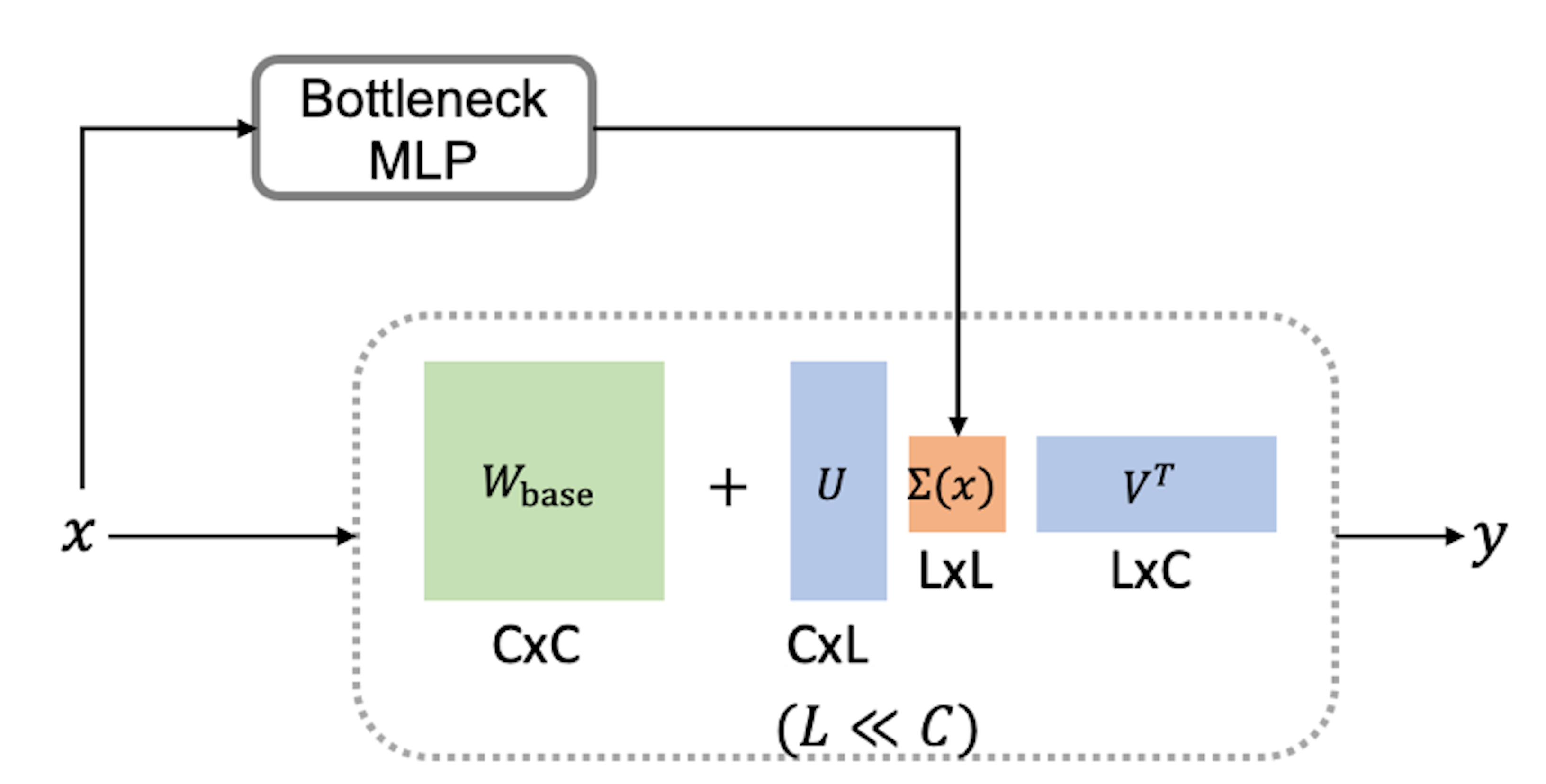}
\caption{TARP with dynamic decomposition (only the additive $\Phi_{2}^l$ is depicted for simplicity).}
\label{fig:dynamic}
\end{figure}

In addition, we propose a novel decomposition inspired by the self-attention mechanism, which aggregate features using input-dependent attention weights. This can be regarded as a function with input-dependent parameters $y = f_{\theta(x)}(x)$. 
Similarly, the optimal reparameterization may vary with different input values. The computation of a reparameterized layer in the PLM becomes
${y}=f_{\theta({x})}({x})=[\Phi_1^l(x)\odot W_0^l + \Phi_2^l(x)]x$
where TARP parameters $\Phi_{j}^l(x)$ are modeled by a \textbf{dynamic low-rank decomposition} (\Cref{fig:dynamic}):
\begin{align}
\Phi_{j}^l(x) = U_{j}^l\Sigma_{j}^l(x){V_{j}^l}^{T},
\end{align}
The square matrices $\Sigma_{j}^l(x) \in \mathbb{R}^{r\times r}$ are generated by a lightweight multi-layer perceptron (MLP) for different input vectors and $U_j^l,V_j^l$ are the learnable weight matrices with $r\ll\text{min}(C_{\text{in}},C_{\text{out}})$.

We compare popular parameter-efficient transfer schemes and these three  decompositions in \Cref{sec:analysis}.

\subsection{Efficient architecture adaptation}
\label{sec:task-adaptive}

We also propose adapting the {model structure} for each task in a data-driven manner. 
The weights of the \textit{task-specific architectures} are learned in the inner loop, while the \textit{task-aware architecture generator} which produces architecture candidates is learned in the outer loop. We term this approach \textbf{task-adaptive model structure (TAMS)} learning.

We first represent each task $\mathcal{T}_i$ with an embedding vector $z_i$ based on the task training set $D_{i}^{\text{train}}$. An embedding module $\mathcal{E}$ computes the task representation by aggregating features of all training data:
\setlength{\abovedisplayskip}{0pt}
\begin{equation}
    z_i = \mathcal{E}(D_{i}^{\text{train}}) = \frac{\sum_{(x,y)\in D_{i}^{\text{train}}} \text{Embed}(x)}{|D_{i}^{\text{train}}|},
\end{equation}
where $\text{Embed}(x)$ are intermediate representations produced by the PLM. For encoder-decoder models (Transformer), we take Embed to be the encoder; for encoder-only or decoder-only PLMs (BERT, GPT-2), we use the token embedding layer.

Inspired by DARTS \citep{liu19darts},
we define the possible sublayer structures by a search space expressed as a directed acyclic graph (DAG),
where each directed edge corresponds to a set of candidate operations $\mathcal{O}$. The task architecture is represented by a set of parameters $\alpha_i$ that encode the structure, where $\alpha_i\in\mathbb{R}^{{E \times |\mathcal{O}|}}$ ($E$ is the number of edges, and $|\mathcal{O}|$ is the number of operations).
In our proposed TAMS approach we also introduce a controller $\mathcal{A}$ to generate these task-specific architecture parameters, as a function of the task embedding vector $\alpha_i = \mathcal{A}(z_i)$. 
The probability of choosing operation $m$ in edge $n$ is given by
$P_n(m)=\text{softmax}_{m\in\mathcal{O}}(\alpha_i[n,m])$.
In meta-testing, the discrete architecture is obtained by taking the argmax.
Since argmax is non-differentiable, we use the straight-through Gumbel-Softmax estimator to backpropagate gradients for optimizing the architecture controller during meta-training.

In TAMS, all possible architectures are initialized as part of the meta-parameters $\tilde{w}$ based on weight-sharing  \citep{pham2018efficient}, i.e., architecture $\alpha_i$'s weights $\tilde{w}(\alpha_i)$ are selected from the meta-parameters.
After the reparameterization steps in TARP and the architecture generation steps in TAMS, 
our inner loop optimization takes parameters $(\Phi, \tilde{w}(\alpha_i))$ and performs a small number of gradient steps $T_\text{in}$ on the task training set to give $(\Phi_i, \tilde{w}_i)$.
In the outer loop optimization, we thus have to simultaneously optimize the architecture controller to perform architecture search, as well as the parameter initialization. This is in contrast to MAML which just optimizes the parameter initialization in the outer loop.
The meta-loss becomes:
$\underset{\mathcal{W}}{\text{min}}\sum_{\mathcal{T}_i\sim p(\mathcal{T})}\mathcal{L}_{D_{i}^{\text{test}}}(f_{\Theta_{\text{LM}}\cup{\Phi_i}\cup\tilde{w}_i})$,
where the tuple $\mathcal{W}$ contains the base PLM's weights $\Theta_{\text{LM}}$, the low-rank reparameterization weights $\Phi$, 
the architecture controller $\mathcal{A}$, and the weight-sharing meta-parameters $\tilde{w}$.

In summary, our contributions with the TAMS framework are that 
(1) 
it meta-learns a task-aware controller by training on the task distribution and then generalizes to new tasks by automatically generating an optimized architecture $\alpha_i$ from the task training data, and
(2) it optimizes the controller and parameter initialization (shared by all tasks) simultaneously under a unified meta-learning objective. 
This is in contrast to DARTS, which performs a separate search and architecture parameter optimization for each task independently.

We summarize our net method with TARP and TAMS in Algorithm \ref{algorithm: overview}.

\begin{algorithm}[t]
\scriptsize
    \textbf{Require}: pretraining dataset $D^{\text{pre}}$; meta-training dataset of tasks $D^{\text{meta}}$; base model (LM) $f_{{\Theta}}$; TARP weights ${\Phi}$; embedding module $\mathcal{E}$ in TAMS; architecture controller $\mathcal{A}$ in TAMS; meta-parameters $\tilde{{w}}$ in TAMS; 
    inner-/outer-loop learning rate $\eta_{\text{in}}$/$\eta_{\text{out}}$; meta-training iterations $T_{\text{meta}}$; inner-loop iterations $T_\text{in}$; meta-batch size $B$\;
    \tcc{Pretraining phase in MLtD}
    {Pretrain the base LM’s weights ${\Theta}\rightarrow{\Theta}_{\text{LM}}$ on $D^{\text{pre}}$}\;
    \tcp{\textcolor{darkpastelred}{In contrast, MAML runs on random initialization.}}
    \tcc{Meta-training phase in MLtD}
    \For{each meta-iteration $t\in[T_{\text{meta}}]$}
    {
        Sample a batch of tasks $\{\mathcal{T}_i\}_{i=1}^B$ from $D^{\text{meta}}$\;
        $\mathcal{L}_{\text{meta}\_\text{loss}} = 0$\;
        \For{$\mathcal{T}_i\in\{\mathcal{T}_i\}_{i=1}^B$}
        {
        {Initialize $\Phi_i = \Phi$}\;
        {Reparameterize $\Theta_{\text{LM}}$ with $\Phi_i$ (Eq.1,3)}\;
        {Expand task-specific sublayers by TAMS-generated architecture $\alpha_i\hspace{-0.03in}=\hspace{-0.03in}\mathcal{A}(\mathcal{E}(D_i^{\text{train}}))$}\;
        {Initialize sublayer's weights: $\tilde{{w}}_i = \tilde{{w}}(\alpha_i)$}\;
        \tcp{\textcolor{darkpastelred}{In contrast, MAML does not adapt model architecture to a task.}}
        \For{$T_\text{in}$ iterations}
        {
            $\mathcal{L}_{\text{inner}}=\mathcal{L}_{D_i^{\text{train}}}\big(f_{{\Theta}_{\text{LM}}\cup{\Phi_i}\cup{\tilde{{w}}_i}}\big)$\;
            $\big({\Phi}_i,\tilde{{w}}_i\big)$~--=~$  \eta_{\text{in}}\nabla_{({\Phi}_i,\tilde{{w}}_i)}\mathcal{L}_{\text{inner}}$\;
            \tcp{\textcolor{darkpastelred}{In contrast, MAML updates all parameters, but we only update a small number in the inner loop.}}
        }
        Evaluate on $D_i^{\text{test}}$:\\
        $\mathcal{L}_{\text{meta}\_\text{loss}}$~+=~$\mathcal{L}_{D_i^{\text{test}}}\big(f_{{\Theta}_{\text{LM}}\cup{\Phi}_i\cup{\tilde{w}_i}}\big)$\;
    }
    Perform outer-loop optimization:
    $\big( {\Theta_{\text{LM}}}, {\Phi}, \mathcal{A}, \tilde{{w}} \big)$~--=~$ \eta_{out}\nabla_{({\Theta_{\text{LM}}}, {\Phi}, \mathcal{A}, \tilde{{w}})}\mathcal{L}_{\text{meta}\_\text{loss}}$\;
    }
    \textbf{Return}: meta-trained PLM with learned $({\Theta_{\text{LM}}}, {\Phi}, \mathcal{A}, \tilde{{w}})$\;
\caption{Meta-Learning the Difference (MLtD) with TARP and TAMS.}
\label{algorithm: overview}
\end{algorithm}

\section{Main results}
\label{sec:experiments}
To demonstrate the overall benefit of our method, we
compare our results to other approaches on generative adaptation tasks in the few-shot (dialogue personalization), low-resource (abstractive summarization), and medium-resource (multi-domain language modeling) regimes.
In \Cref{sec:analysis} we perform some analyses and also compare TARP by itself to previous parameter-efficient works.

\subsection{Implementation}
\label{sec:tams-details}

All of our experiments ran using PyTorch on single machines with NVIDIA V100 GPUs. 
See per-task hyperparameters in \Cref{sec:hyperparams}.

\paragraph{TARP decomposition.} We apply task-adaptive reparameterization (TARP) to the pretrained self-attention and feed-forward network (FFN) blocks. In \Cref{sec:efficiency-exps} we conclude that TARP with dynamic decomposition outperforms other parameter-efficient transfer methods; TARP will always be of this form for our main experiments, with rank $r \le 32$.

\paragraph{TAMS details.}
We apply TAMS to expand the FFN block, so while the shared (in structure) sublayers capture the commonalities among tasks, the new searched sublayers can capture task-specific structure.
Our search DAG contains two input nodes that project the inputs to a low-dimensional space, one output node that projects the intermediate representation back to the original dimension, and three intermediate nodes. Candidate operations for each edge are \{linear, conv-3$\times$1, conv-5$\times$1, gated linear unit (GLU), zeroize, and skip connection\}; see code for definitions.
All the candidates operate on a reduced feature dimension to ensure the parameter efficiency of the search cell.
Our controller $\mathcal{A}$ is a two-layer MLP. The first fully-connected layer has 128 output neurons, and the second layer has $E \times |\mathcal{O}|$ neurons (see \Cref{sec:task-adaptive} for notation). We apply ReLU after the first layer and softmax the final output.

\subsection{Few-shot dialogue personalization}
\label{sec:persona-chat}

Persona-Chat \citep{zhang-etal-2018-personalizing} is a dialogue generation benchmark with 1137/99/100 personas for training/validation/testing. We follow recent work \citep{madotto-etal-2019-personalizing, song-etal-2020-learning} and regard learning a dialogue model for each persona as a few-shot meta-learning task. On average, each persona has 8.3 unique dialogues, 6-8 turns per dialogue, and 15 words per turn.
Following these works, we use a standard Transformer model with pretrained GLoVe embeddings and separate the dialogues by their persona description into meta-training/-validation/-testing using \citet{madotto-etal-2019-personalizing}'s splits and code\footnote{\url{https://github.com/HLTCHKUST/PAML}}.

\paragraph{Baselines.} The following are from previous works. \textbf{Pretrain} denotes a multitask dialogue model trained on labeled data from all meta-training tasks. \textbf{MAML} meta-trains the Transformer model from scratch \citep{madotto-etal-2019-personalizing}, and \textbf{CMAML} \citep{song-etal-2020-learning} additionally applies a pruning algorithm to customize the model structures for different tasks. \textbf{+Finetune} corresponds to finetuning on each testing task.
Finally, \textbf{Pretrain+Persona} is a partial oracle for reference only, where the persona description is available.

\begin{table}[t]
\centering
\footnotesize
\captionsetup{font=small}
\captionsetup[sub]{font=small}
\begin{tabular}{@{}lcc@{}}
    \toprule
    \textbf{Method} & \textbf{PPL} & \textbf{BLEU} \\
    \midrule
    Pretrain (multitask)$^*$ & 36.75 & 0.64 \\
    Pretrain+Finetune$^*$ & 33.14 & 0.90 \\
    MAML+Finetune$^*$ & 40.34 & 0.74 \\
    CMAML+Finetune$^*$ & 36.30 & 0.89 \\
    \textit{Pretrain+Persona}$^*$ & \textit{30.42} & \textit{1.00} \\
    \midrule
    Pretrain+MAML+Finetune & 32.54 & 0.97 \\
    MLtD (TARP only) & 32.15 & 0.99 \\
    MLtD & \textbf{28.14} & \textbf{1.20} \\\bottomrule
\end{tabular}
\caption{Comparison of test perplexity (PPL; lower is better) and BLEU (higher is better) for few-shot dialogue generation on Persona-Chat dataset. $^*$: published results from \citet{madotto-etal-2019-personalizing, song-etal-2020-learning}; the rest are ours.}
\label{tab:meta_learning}
\end{table}

\begingroup
\setlength{\tabcolsep}{6pt}
\begin{table*}[t]
\centering
\footnotesize
\captionsetup{font=small}
\captionsetup[sub]{font=small}
\begin{tabular}{@{}l c c c c c c | c @{}} \toprule
	\textbf{Method} & \textbf{Dialog} &	\textbf{Email} & \textbf{Movie} & \textbf{Debate} & \textbf{Social} & \textbf{Science} & \textbf{Avg.} \\\midrule
Baseline (full finetuning)$^*$ & 39.95 &	24.71 &	25.13 &	24.48 &	21.76 & 72.76 &	34.80 \\
DAPT (Domain-Adaptive Pre-Training)$^*$ &	41.22 &	26.50 &	24.25 &	26.71 &	22.95 & 71.88 & 35.59 \\
TAPT (Task-Adaptive Pre-Training)$^*$ &	40.15 &	25.30 &	25.27 &	24.59 &	22.81 & 73.08 &	35.20 \\
SDPT (Supervised Domain Pre-Training)$^*$ &	42.84 &	25.16 &	25.45 &	25.61 &	22.43 & 73.09 &	35.76 \\
\midrule
MLtD & \textbf{44.81} & 25.30 &	\textbf{26.83} &	\textbf{26.88} &	\textbf{24.40} & \textbf{74.03} &	\textbf{37.04} \\
\;\;\;(TARP only) &	42.88 &	\textbf{26.92} &	25.98 &	25.95 &	23.34 & 73.69 & 36.46 \\
\;\;\;(TARP only, no meta-learning) & 40.39 &	23.20 &	25.81 &	26.67 &	21.46 & 73.20 &	35.12 \\
\;\;\;(TARP only, multitask pretraining instead) & 41.82 & 25.41 & 26.17 & 25.70 & 22.54 & 73.50 & 35.85 \\
\bottomrule
\end{tabular}
\caption{ROUGE F1s from multi-domain adaptation for abstractive summarization on AdaptSum (higher is better). All methods are initialized with pretrained BART and finetuned on the labeled task training set of each domain at the end. $^*$: published results from \citet{yu-etal-2021-adaptsum}, using DAPT and TAPT methods from \citet{gururangan-etal-2020-dont}; the rest are ours.}
\label{tab:adaptsum_abs}
\end{table*}
\endgroup

\paragraph{Results (Table \ref{tab:meta_learning}).}
We include the same evaluation metrics from previous works (perplexity, BLEU score). 
Training MAML from scratch yields worse results than the Pretrain model. However, when MAML is initialized from the multitask model (\textbf{Pretrain+MAML+Finetune}), the result already outperforms previous work. 
Note that the \textit{same labeled data} is used for both Pretrain and MAML, suggesting that
meta-learning benefits from the more robust initialization that pretraining provides to improve task-specific few-shot adaptation (also see analysis in \Cref{sec:pre-vs-meta}).

Moreover, we see further improvements by ``meta-learning the difference'' (MLtD). By using TARP for MAML's inner loop adaptation (\textbf{MLtD, TARP only}), we attain equivalent or better results and faster training time while only updating a small amount of task-specific parameters (\Cref{sec:efficiency-exps}). This indicates that our method helps mitigate overfitting to low-resource tasks. 
Finally, by incorporating TAMS (\textbf{MLtD}), we use the full framework and achieve the best performance, suggesting the task-adapted model structure gives better architectures for  personas.
In this regard, \textbf{CMAML} lags behind MLtD as well.
We conjecture this is because it uses a pruning algorithm to
``customize'' the model with different weight masks, which may not generate enough model diversity for diverse tasks as the architectural inductive bias remains the same.

\begingroup
\setlength{\tabcolsep}{8.5pt}
\begin{table*}[ht!]
\centering
\footnotesize
\captionsetup{font=small}
\captionsetup[sub]{font=small}
\begin{tabular}{@{}l c c c c c c | c @{}} \toprule
	\textbf{Method} & \textbf{Dialog} &	\textbf{Email} & \textbf{Movie} & \textbf{Debate} & \textbf{Social} & \textbf{Science} & \textbf{Avg.} \\\midrule
	Baseline (full finetuning) & 31.95 &	31.57 &	42.25 &	34.38 &	33.02 &	28.82 &	33.67 \\
	\;\;\;Zero-shot (no finetuning) &	37.26 &	38.45 &	49.46 &	41.38 &	37.13 &	34.20 &	39.65 \\
	DAPT$^\dagger$ &	35.15 &	\textbf{16.04} &	43.12 &	33.83 &	27.15 &	18.96 &	29.04 \\
	\midrule
	MLtD & 29.66 &	16.93 &	\textbf{35.38} &	\textbf{30.61} &	\textbf{19.78} &	\textbf{17.06} & \textbf{24.90} \\
	\;\;\;(TARP only) &	\textbf{28.63} &	18.67 &	39.73 &	32.70 &	26.93 &	20.39 &	27.84 \\
	\;\;\;(TARP only, no meta-learning) & 31.66 &	31.59 &	41.78 &	33.18 &	32.78 &	28.20 &	33.19 \\
	\bottomrule
\end{tabular}
\caption{Test perplexities from multi-domain language modeling adaptation on AdaptSum (lower is better). All methods are initialized with pretrained GPT-2 medium and finetuned on the labeled domain set at the end. $^\dagger$: our re-implementation of \citet{gururangan-etal-2020-dont}.}
\label{tab:adaptsum_lm}
\end{table*}
\endgroup

\subsection{Low-resource abstractive summarization}
\label{sec:adapt-sum}

AdaptSum \citep{yu-etal-2021-adaptsum} is a new multi-domain dataset used to evaluate domain adaptation schemes for abstractive summarization. It consists of six diverse target domains ranging from movie reviews to scientific abstracts. Each domain has a low-resource task corpus and a larger unlabeled text corpus as well (list and statistics in \Cref{tab:adaptsum_stats}) that is used to evaluate domain- and task-adaptive pretraining (DAPT/TAPT; \citealp{gururangan-etal-2020-dont}).
We use pretrained BART \citep{lewis-etal-2020-bart} and finetune to each low-resource task corpus as in \citet{yu-etal-2021-adaptsum}, whose code\footnote{\url{https://github.com/TysonYu/AdaptSum}} we extend.

\begin{table}[ht!]
\centering
\footnotesize
\captionsetup{font=small}
\captionsetup[sub]{font=small}
\begin{tabular}{@{}l | c | c c c @{}}\toprule
    \multirow{3}{*}{\textbf{Domain}} & \multicolumn{4}{c}{\textbf{\# of tokens}} \\
    & \multirow{2}{*}{\textbf{Text only}} & \multicolumn{3}{c}{\textbf{Task corpus}} \\
    & & \textbf{train} & \textbf{val} & \textbf{test} \\ 
    \midrule
    Dialog & 44.96M & 27K & 74K & 75K \\
    Email  & 117.54M & 37K & 243K & 237K \\
    Movie review & 11.36M & 633K & 1056K & 6193K \\
    Debate & 122.99M & 59K & 188K & 197K \\
    Social media & 153.30M & 68K & 229K & 229K \\
    Science & 41.73M & 63K & 221K & 314K \\
    \bottomrule
\end{tabular}
\caption{Data sizes for AdaptSum \citep{yu-etal-2021-adaptsum} across the six domains, for both the text-only domain-related corpus and the low-resource task corpus.}
\label{tab:adaptsum_stats}
\end{table}

\paragraph{Baselines.} 
\textbf{DAPT} continues pretraining with BART's self-supervised objective using the unlabeled domain corpus. \textbf{TAPT} continues pretraining with the set of unlabeled documents found in the target summarization task. 
\textbf{SDPT} uses the XSum dataset in the News domain to further pretrain BART with a supervised training objective using document-summary pairs before finetuning.

\paragraph{Results (\Cref{tab:adaptsum_abs}).} We find that \textbf{MLtD}, even without architecture search (\textbf{TARP only}), outperforms DAPT, TAPT, and SDPT.
These methods use in-domain/-task knowledge and the standard pretraining objective to help  adaptation to the target task, while our method considers cross-domain knowledge via the meta-learning objective, sampling meta-training tasks from multiple domain corpora to train the model.
Moreover, the use of meta-learning as preparation outperforms multitask pretraining (\textbf{TARP only, multitask pretraining instead}), signifying that mere exposure to the cross-domain data may not be enough and using a meta-learning objective to explicitly optimize for the lightweight adaptation is beneficial.
Finally, we see that without meta-learning or multitasking (\textbf{TARP only, no meta-learning}) our performance is also better than the baseline. 
This demonstrates the effectiveness of the lightweight TARP adaptation, which matches the performance of full finetuning while only updating less than 5\% of parameters.

\subsection{Multi-domain language modeling}

Though the text corpora in AdaptSum were originally included to evaluate DAPT, we also use them to evaluate our methods on multi-domain language modeling. As this is a novel benchmark, to demonstrate fast adaptation we take $T_{\text{in}} = 1$.

\paragraph{Baselines.} We start with pretrained GPT-2 medium (345M) \citep{radford19gpt2} with input sequence length 512 using the Transformers library \citep{huggingface}. Finetuning is performed on the training documents of the task corpus, and we evaluate perplexity on the test documents of the task corpus. 
The only exception to finetuning is \textbf{Zero-shot} which evaluates the pretrained GPT-2 model directly.
\textbf{DAPT} continues pretraining of GPT-2 with the language modeling objective on the unlabeled domain corpus before finetuning.

\paragraph{Results (\Cref{tab:adaptsum_lm}).} Our findings in summarization also hold for the unconditional causal language modeling task. Namely, we see equal or better performance of TARP vs.\ full finetuning and that meta-learning plays a significant role in the task adaptation quality.
In contrast to summarization with BART (\Cref{sec:adapt-sum}) but similar to Persona-Chat with Transformer (\Cref{sec:persona-chat}), we see that TAMS leads to noticeable improvements. 
We explain why this may be the case and present the TAMS-learnt sublayer modules in \Cref{sec:tams-analysis}.

\section{Analysis}
\label{sec:analysis}

\begingroup
\setlength{\tabcolsep}{6pt}
\begin{table*}[t]
\centering
\footnotesize
\captionsetup{font=small}
\captionsetup[sub]{font=small}
\begin{tabular}{l | c | c c c c c | c | c}\toprule
    \multirow{2}{*}{\textbf{Method}} & \textbf{Params.} & \multicolumn{5}{c|}{\textbf{E2E}} & \textbf{DART} & \textbf{WebNLG}  \\
    & \textbf{per task} & \textbf{BLEU} & \textbf{NIST} &	\textbf{METEOR} & \textbf{ROUGE-L} & \textbf{CIDEr} & \textbf{BLEU} & \textbf{BLEU} \\\midrule
    Finetuning (full)$^*$ & 100\% & 68.2 & 8.62 & 46.2 & 71.0 & 2.47 & 46.0 & 47.6 \\
    FT-Top2$^*$ & 7.1\% & 68.1 & 8.59 & 46.0 & 70.8 & 2.41 & 38.1 & 33.5 \\
    BitFit$^*$ & 0.1\% & 67.2 & 8.63 & 45.1 & 69.3 & 2.32 & 43.3 & 50.5 \\
    Adapter$^*$ & 3.2\% & 68.9 & 8.71 & 46.1 & 71.3 & 2.47 & 45.4 & 54.0 \\
    \midrule
    Bilinear TARP & 2.4\% & 68.8 & 8.75 & 46.1 & 70.8 & 2.43 & 46.7 & 54.0 \\
    Kronecker TARP & 2.4\% & 68.2 & 8.73 & 45.2 & 69.4 & 2.36 & 45.6 & 53.1 \\
    Dynamic TARP & 1\% & \textbf{69.7$_{\pm0.1}$} & \textbf{8.78$_{\pm0.02}$} & \textbf{46.9$_{\pm0.2}$} & \textbf{72.1$_{\pm0.1}$} & \textbf{2.51$_{\pm0.01}$} & \textbf{47.9$_{\pm0.2}$} & \textbf{55.3$_{\pm0.1}$} \\
    {w/ matrix mult.} & {1\%} & {68.3} & {8.64} & {46.4} & {71.1} & {2.47} &  46.5 & 53.2 \\\bottomrule
\end{tabular}
\caption{Comparison with other adaptation methods for natural language generation (E2E, DART, WebNLG) on GPT-2 medium. $^*$: published results from \citet{houlsby19adapters,zaken21bitfit}; the rest are ours. For our dynamic TARP, we provide the 95\% confidence interval over 5 runs.}
\label{tab:sota_adaptation}
\end{table*}
\endgroup

\begingroup
\setlength{\tabcolsep}{4pt}
\begin{table*}[h]
\centering
\footnotesize
\captionsetup{font=small}
\captionsetup[sub]{font=small}
\begin{tabular}{@{}l | c | c c c c c c c c | c@{}}\toprule
    \multirow{2}{*}{\textbf{Method}} & \textbf{Params.} & \textbf{CoLA} & \textbf{MRPC} & \textbf{STS-B} & \textbf{RTE} & \textbf{SST-2} & \textbf{MNLI} & \textbf{QNLI} & \textbf{QQP} & \textbf{Avg.} \\
     & \textbf{per task} & \textbf{Matt.\ corr} & \textbf{Acc.} & \textbf{Pear.\ corr} & \textbf{Acc.} & \textbf{Acc.} & \textbf{Acc.} & \textbf{Acc.} & \textbf{Acc.} & \\ \midrule
     Finetuning (full)$^*$ & 100\% & 63.6 & {90.2} & 91.2 & 78.7 & 94.8 & \textbf{87.6} &	92.8 &	91.9 &	86.4 \\
     Masking$^*$ & 3.1\% & 60.3 & 88.5 & - & 69.2 & 94.5 & - & 92.4 & - & - \\
     BitFit$^*$ & 0.1\% & 61.8 & \textbf{92.0} & 90.8 & 77.8 & 93.7 & 85.2 & 91.3 & 84.5 & 84.6 \\
     LoRA$^\dagger$ & 1\% & 62.2 & 89.6 & 90.6 & 72.9 & 94.8 & 87.4 & 92.4 & 92.3 & 85.3 \\
     AdapterFusion$^*$ & 1\% & - & 89.7 & - & 78.8 & 93.7 & 86.2 & - & 90.3 & -  \\
     MAM Adapter$^*$ & 0.5\% & 59.2 &	88.5 &	90.6 &	74.3 &	94.2 &	87.4 &	92.6 &	90.2 &	84.6 \\
     MAM Adapter$^\dagger$ & 1\% & 63.7 &	89.4 &	90.6 &	76.6 &	94.4 &	\textbf{87.6} &	92.8 &	90.9 &	85.7 \\
     \midrule
     Dynamic TARP & 1\% & \textbf{64.5$_{\pm.7}$} &	89.9$_{\pm.3}$	& \textbf{91.3$_{\pm.2}$} &	\textbf{79.1$_{\pm.6}$} &	\textbf{94.9$_{\pm.1}$} &	\textbf{87.6$_{\pm.2}$} &	\textbf{93.3$_{\pm.1}$} &	\textbf{93.1$_{\pm.1}$} &	\textbf{86.7} \\
    \bottomrule
\end{tabular}
\caption{Comparison with other adaptation methods on low-resource GLUE tasks.
We report single-task results (including our dynamic TARP) of adapting RoBERTa base on the training set of each task only.
$^*$: published results from \citet{liu19roberta,zhao-etal-2020-masking,zaken21bitfit,pfeiffer-etal-2021-adapterfusion,he2022towards}; $^\dagger$: recreated using \citet{he2022towards}'s implementation; the rest are ours.
For our dynamic TARP, we provide the 95\% confidence interval over 5 runs.}
\label{tab:sota_adaptation_glue}
\end{table*}
\endgroup

\subsection{Pretraining improves meta-learning}
\label{sec:pre-vs-meta}

\begin{figure}[t]
\small
\captionsetup{font=small}
\captionsetup[sub]{font=small}
\centering
\begin{subfigure}[b]{0.49\linewidth}
  \includegraphics[scale=0.145]{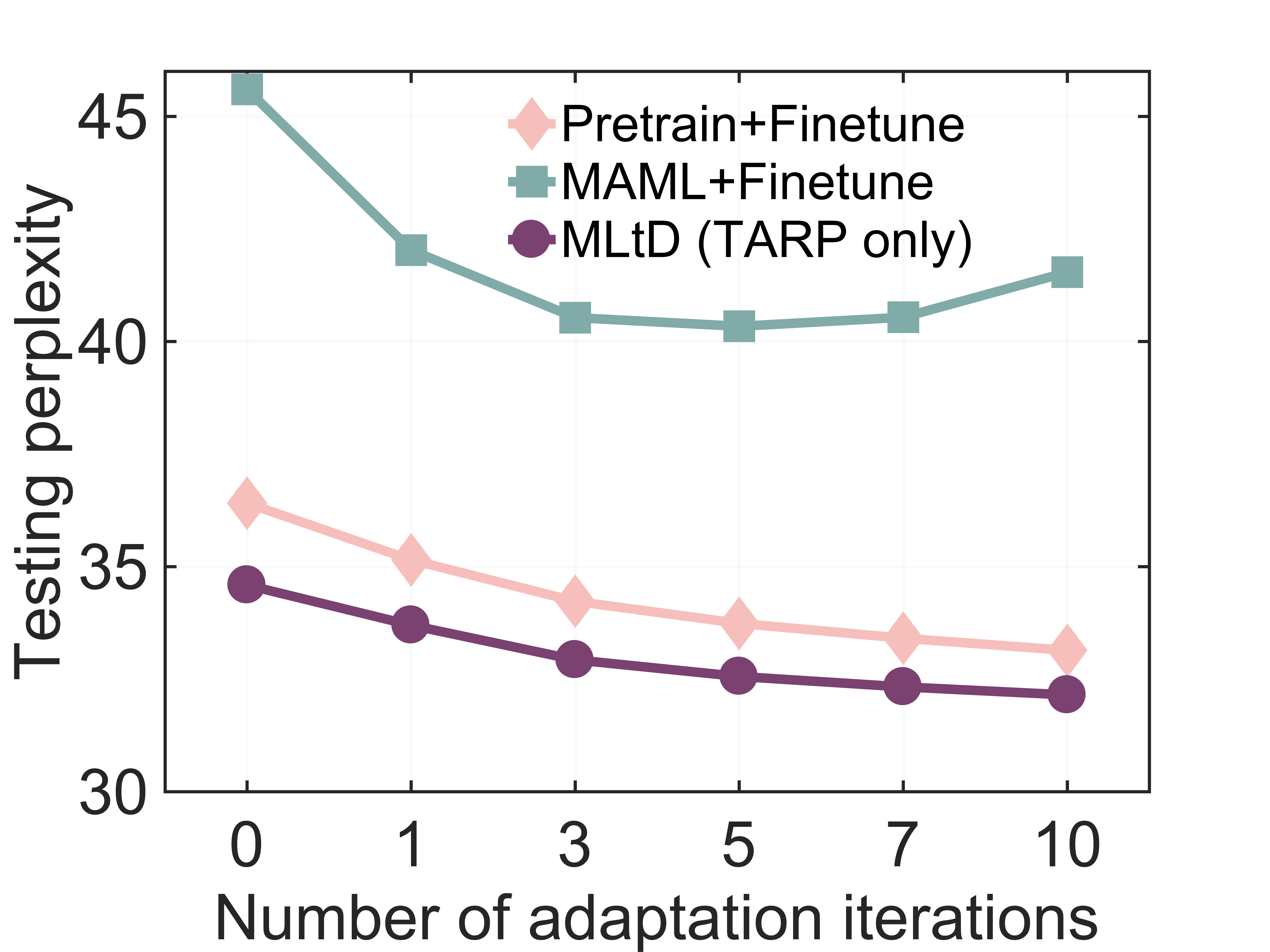}
  \label{fig:ppl_vs_adapt}
  \vspace{-0.15in}
\end{subfigure}~
\begin{subfigure}[b]{0.49\linewidth}
\centering
  \includegraphics[scale=0.145]{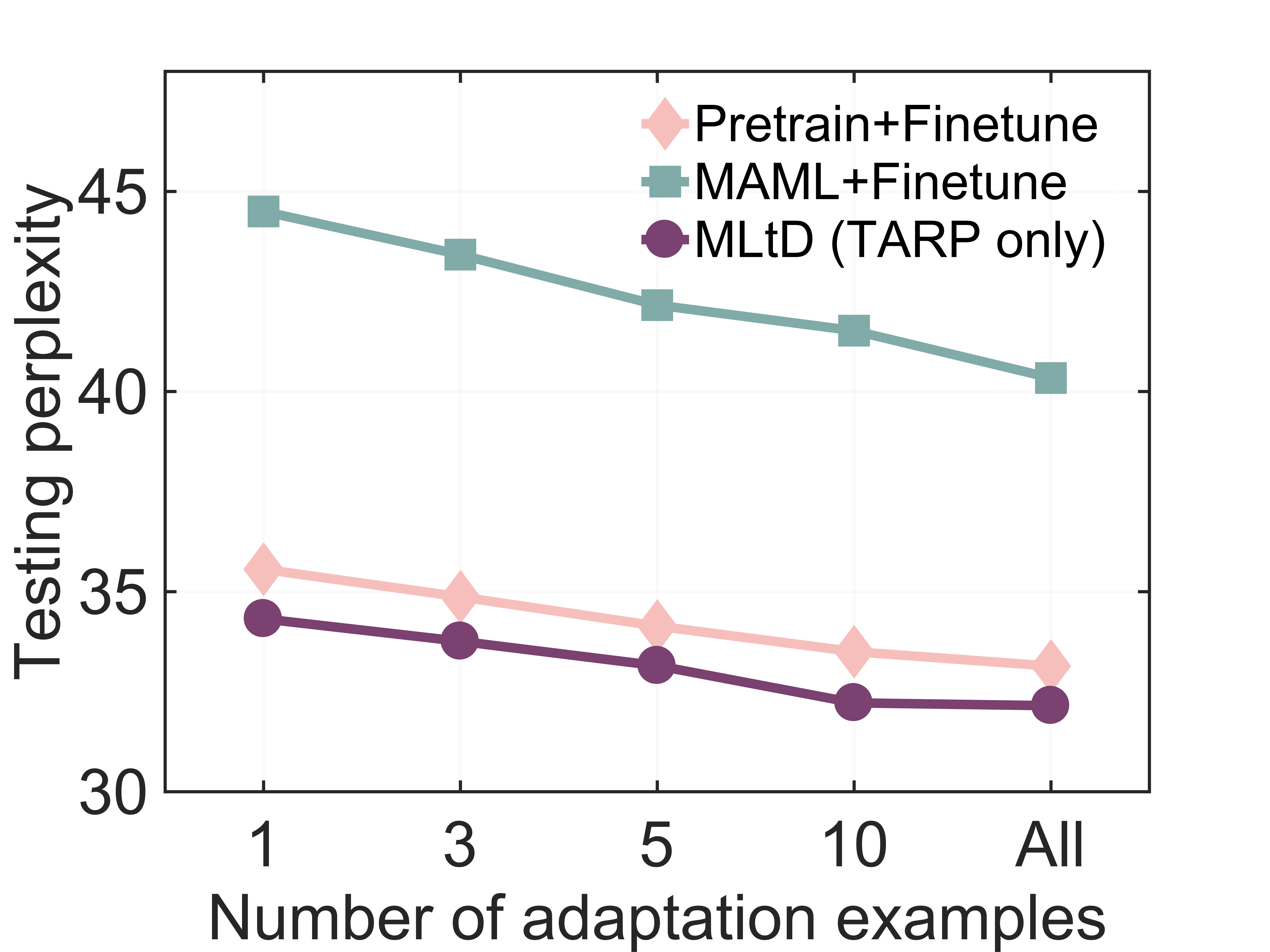}
  \label{fig:ppl_vs_dia}
  \vspace{-0.15in}
\end{subfigure}
\caption{Perplexities on Persona-Chat testing tasks with MLtD (TARP only) versus Pretrain+Finetune and MAML+Finetune. \textbf{Left}: Influence of number of adaptation iterations. \textbf{Right}: Influence of the number of adaptation dialogues.
}
\label{fig:ablation_metalearning}
\end{figure}

We analyze the performance of MLtD on Persona-Chat at meta-testing time (i.e., finetuning then testing on unseen personas) with respect to the number of inner loop steps and training dialogues. In \Cref{fig:ablation_metalearning} (left), we see that original MAML (no pretraining) overfits,
while finetuning the multitask-pretrained model keeps improving.
Moreover, MLtD atop the multitask-pretrained model followed by finetuning continues to improve test perplexity.
In \Cref{fig:ablation_metalearning} (right), we fix the finetuning steps and vary the number of training dialogues used in finetuning. 
Using more dialogues improves perplexity for all three methods, with MLtD still leading over full MAML and direct finetuning after pretraining.
The takeaway from these results is that applying MAML on a pretrained model prevents overfitting and promotes better generalizability from meta-learning.

\subsection{Dynamic TARP versus alternatives}
\label{sec:efficiency-exps}

We benchmark our dynamic low-rank reparameterization on a variety of NLP models and tasks. 
To show that dynamic TARP individually improves on full finetuning and other parameter-efficient adaptation methods, we report single-task results here.
For generative tasks, we use pretrained GPT-2 medium on
natural language generation datasets:
we specifically evaluate on E2E \citep{novikova-etal-2017-e2e}, which was used by adapter method \citep{lin-etal-2020-exploring}; WebNLG \cite{gardent-etal-2017-webnlg}; and DART \citep{nan-etal-2021-dart}, which was used by LoRA \citep{hu21lora}.
For classification, we use pretrained RoBERTa \citep{liu19roberta} on low-resource GLUE \citep{wang19glue} tasks, which were evaluated on by many recent parameter efficient adaptation methods \citep{houlsby19adapters, zhao-etal-2020-masking, zaken21bitfit, he2022towards}. 
Further dataset and experimental setup details are in \Cref{sec:adaptation-data}. 
In particular, we chose rank $r$ to give similar parameter counts to other approaches; $r=4$ in \Cref{tab:sota_adaptation}, $r=8$ in \Cref{tab:sota_adaptation_glue}.

For generative tasks, we compare with \textbf{finetuning} all layers; \textbf{FT-Top2}, 
which only finetunes the last two layers of the model; \textbf{BitFit} \citep{zaken21bitfit}, which only finetunes the biases; and
\textbf{Adapter tuning} \citep{houlsby19adapters}, which only finetunes the adapter layers inserted after each feed-forward and self-attention sublayer.
As shown in \Cref{tab:sota_adaptation}, TARP methods match or outperform other parameter-efficient methods, while learning task-specific parameters that are $<$3\% of the number of base parameters and keep the base model unchanged.
Among the three TARP variants, we find that \textbf{Dynamic} $>$ \textbf{Bilinear} $>$ \textbf{Kronecker} in terms of performance across generative metrics. 
This suggests that the optimal adaptation to the underlying model weights may vary per token, which dynamic low-rank accounts for.
Moreover, dynamic TARP performs better than an alternative where the $O(n^2)$ Hadamard product in Eq.\eqref{reparameterization} is replaced by  $O(n^3)$ matrix multiplication (\textbf{w/ matrix mult.}).

For classification tasks, we compare with \citet{he2022towards}, which proposed a unified framework connecting several state-of-the-art adaptation methods \citep{houlsby19adapters,hu21lora,li-liang-2021-prefix} and devised an improved method (\textbf{MAM Adapter}).
Our dynamic TARP can only partly be viewed in this unified framework 
as we explore a novel design dimension, i.e., making the modification to the base model dynamic w.r.t.\ input tokens. Moreover, in contrast to additive-only modifications in \citet{he2022towards}, our dynamic TARP applies both multiplicative and additive modifications.
For fair comparisons, we follow past works \citep{liu19roberta,zhao-etal-2020-masking,he2022towards} and set the maximum finetuning epochs to 10 on each task. 
In \Cref{tab:sota_adaptation_glue},
dynamic TARP introduces and trains only 1\% versus the number of original parameters, while achieving comparable results to full finetuning (+0.3 abs.) and outperforming the previous best, MAM adapters (+1.0 abs.).

\subsection{Dynamic TARP outperforms finetuning}

\begin{figure}[t]
\centering
\small
\captionsetup{font=small}
\captionsetup[sub]{font=small}
\begin{subfigure}[b]{0.49\linewidth}
  \includegraphics[scale=0.145]{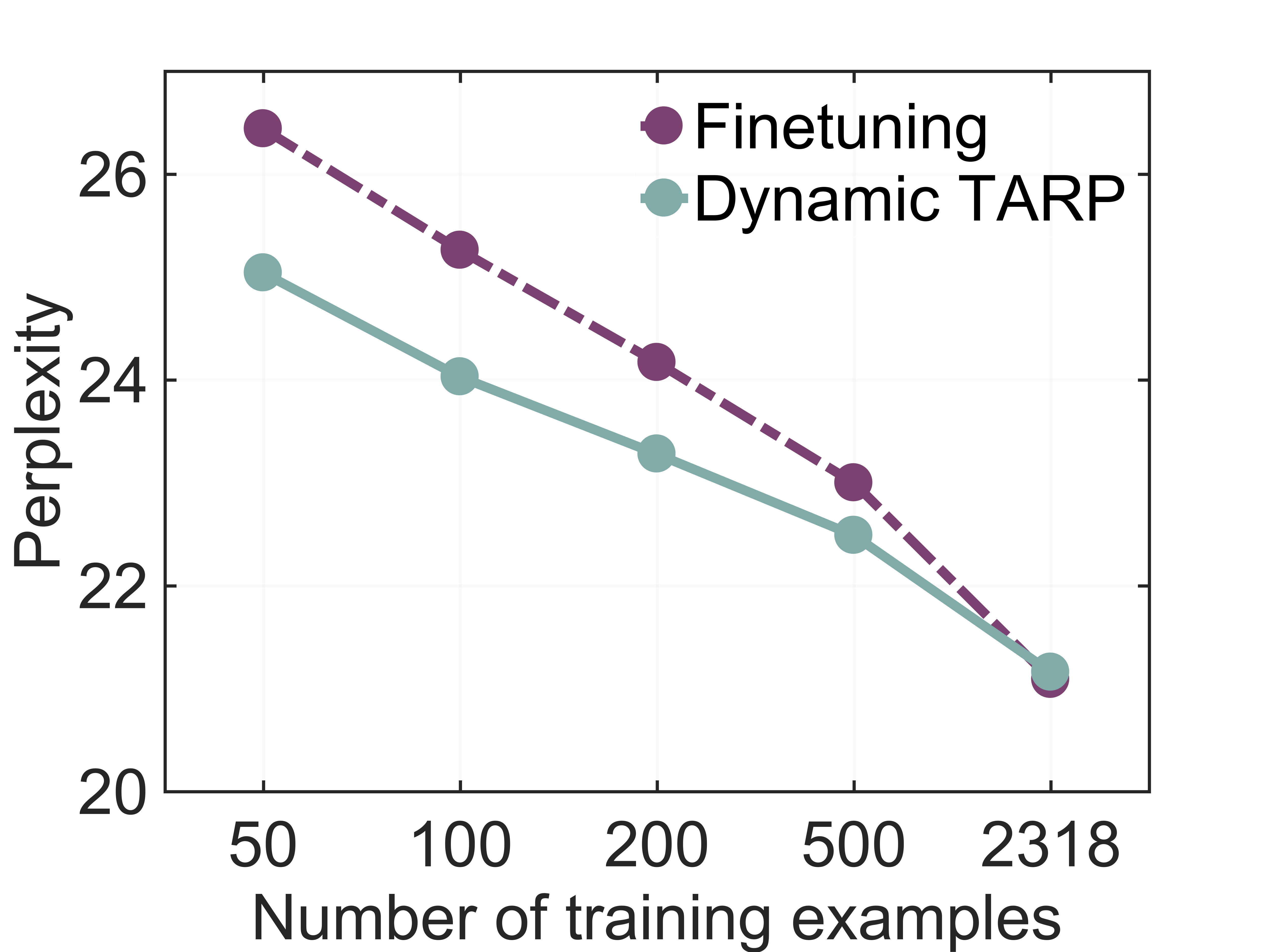}
  \vspace{-0.15in}
\end{subfigure}~
\begin{subfigure}[b]{0.49\linewidth}
\centering
  \includegraphics[scale=0.145]{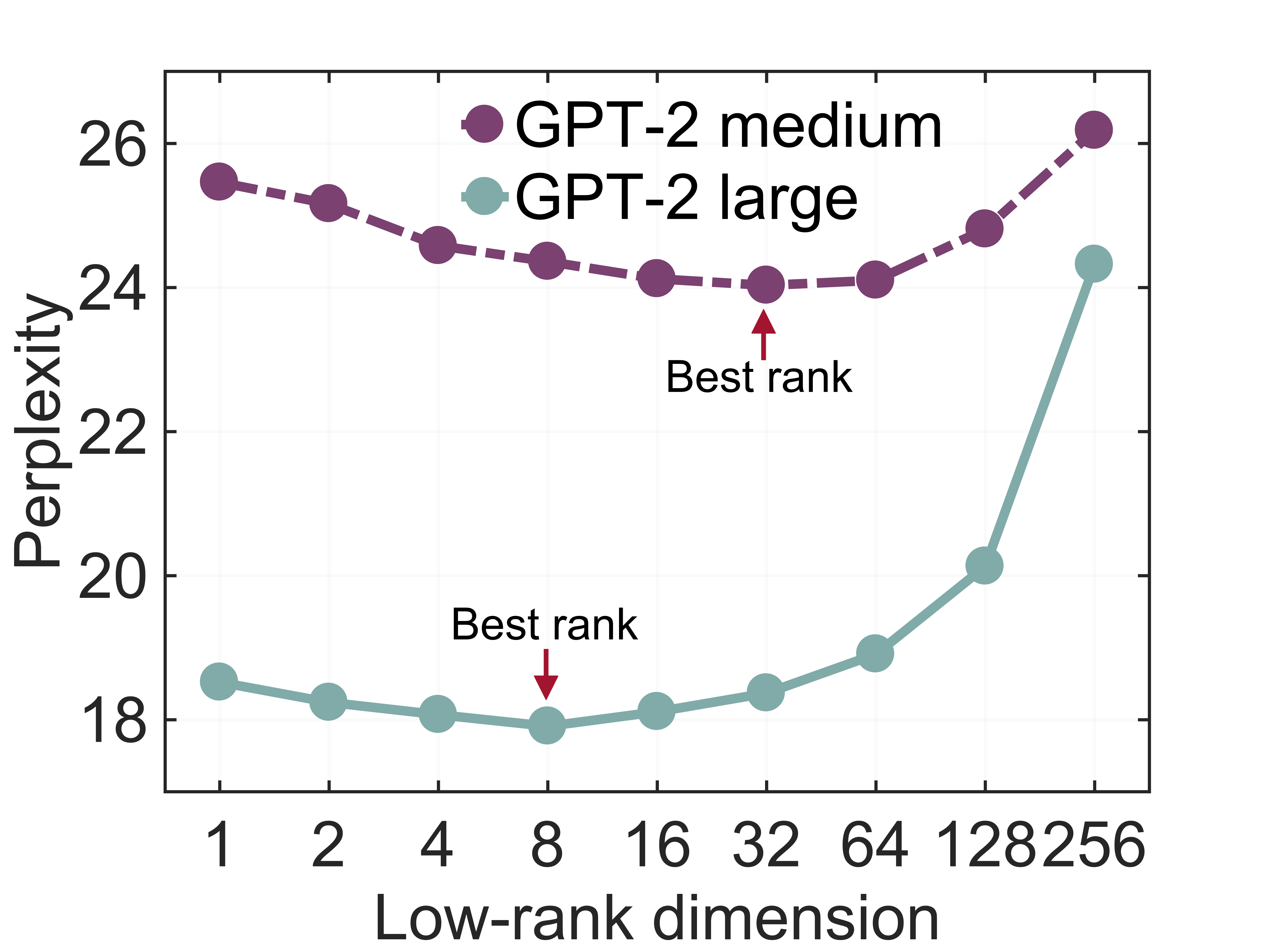}
  \vspace{-0.15in}
\end{subfigure}
\caption{Testing perplexities on WikiText-2 with full finetuning and/or low-rank adaptation with dynamic TARP.
\textbf{Left}: Low-rank adaptation is extremely helpful on low-resource tasks.
\textbf{Right}: Holding the number of training examples fixed, the model adaptation space is optimized by fewer dimensions.
}
\label{fig:lr_ablations}
\end{figure}

Tables \ref{tab:sota_adaptation} and \ref{tab:sota_adaptation_glue} also show that dynamic low-rank reparameterization outperforms finetuning on corresponding evaluating metrics, while being faster as it only adapts a small set of weights.
The training time further improves through utilizing the training data more efficiently. 
In \Cref{fig:lr_ablations} (left) we compare perplexities of our method against finetuning on subsets of WikiText-2 and see that finetuning increasingly underperforms as the number of examples decrease.
To explain this behavior, in \Cref{fig:lr_ablations} (right) we fix the number of training examples to 100 and ablate the rank. Our method performs best with a very small rank value, suggesting that the difference between the pretrained and finetuned weight matrices lies in a lower-dimensional subspace.
This complements \citet{aghajanyan-etal-2021-intrinsic}'s observation that direct adaptation in lower-dimensional spaces can be equally as effective as in the original space. 
Moreover, we find that the larger the model (GPT-2 medium vs.\ the GPT-2 small), the lower the rank value required for the best adaptation performance. 

\subsection{TAMS discovers better architectures}
\label{sec:tams-analysis}

Recent studies have shown that simple modifications to the transformer architecture, such as re-organizing the MHSA and FFN modules \cite{zhao-etal-2021-memory} or adding 1D convolutions to self-attention \cite{so2021primer}, improve the task performance.
Similarly, from our results in \Cref{tab:adaptsum_lm}, adapting the model structure through sub-layer modifications in our meta-learning framework further reduces the testing perplexity compared to MLtD with fixed model structure.
Applying task-aware architecture search (TAMS) on the FFN module incurs less than 5\% additional model parameters compared to the original GPT-2 model, but reduces the perplexity by 3 points on average.

\begin{figure}[t]
\centering
\small
\captionsetup{font=small}
\captionsetup[sub]{font=small}
\includegraphics[width=\linewidth]{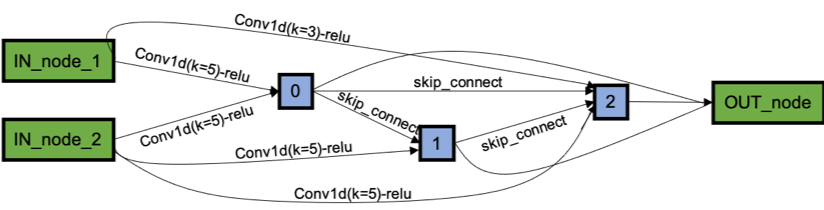}
\caption{Dominant structure of the TAMS-learned sublayers for AdaptSum language modeling.}
\label{fig:searched_FFN}
\end{figure}

A limitation we observe is that the TAMS method tends to produce a dominant architecture (cf.\ \Cref{fig:searched_FFN}) as opposed to one different architecture for each task. We conjecture this may be because our initial task representation strategy has low variance due to averaging across the entire task training data. This may explain why TAMS did not uniformly improve MLtD in all settings. Nevertheless, the perplexity reduction implies that there is still room to optimize the architecture of current LMs without significantly increasing total model size. Thus, we believe task-aware architecture search is a promising direction to continue to invest in the future.

\subsection{Training efficiency of MLtD}
We study training efficiency by comparing the training and finetuning wall-clock time for multi-domain abstractive summarization on AdaptSum. The results are shown in Table \ref{tab:adaptsum_abs_gputime}.

\begingroup
\setlength{\tabcolsep}{2pt}
\begin{table}[t]
\centering
\footnotesize
\captionsetup{font=small}
\captionsetup[sub]{font=small}
\begin{tabular}{l | c c} \toprule
\multirow{2}{*}{\textbf{Method}} & \textbf{Prep.} & \textbf{Finetuning}\\
& \textbf{(hrs.)} & \textbf{(mins.)} \\ \midrule
Baseline (direct finetuning) & -- & 26 \\
SDPT & 64 & 16 \\
DAPT & 208 & 23 \\
TAPT & {8} & 18 \\
\midrule
MLtD & 39 & 9 \\
\;\;\;(TARP only) & 22 & 7 \\
\;\;\;(TARP only, no meta-learning) & -- & 20 \\
\bottomrule
\end{tabular}
\caption{Wall-clock time comparison on AdaptSum during preparation on the meta-training data (Prep.) and during meta-testing (Finetuning) to convergence (early stopping), summed over all domains.
Times were measured on one GPU.
}
\label{tab:adaptsum_abs_gputime}
\end{table}
\endgroup

\begin{figure}[t]
\centering
\captionsetup{font=small}
\captionsetup[sub]{font=small}
\includegraphics[scale=0.2]{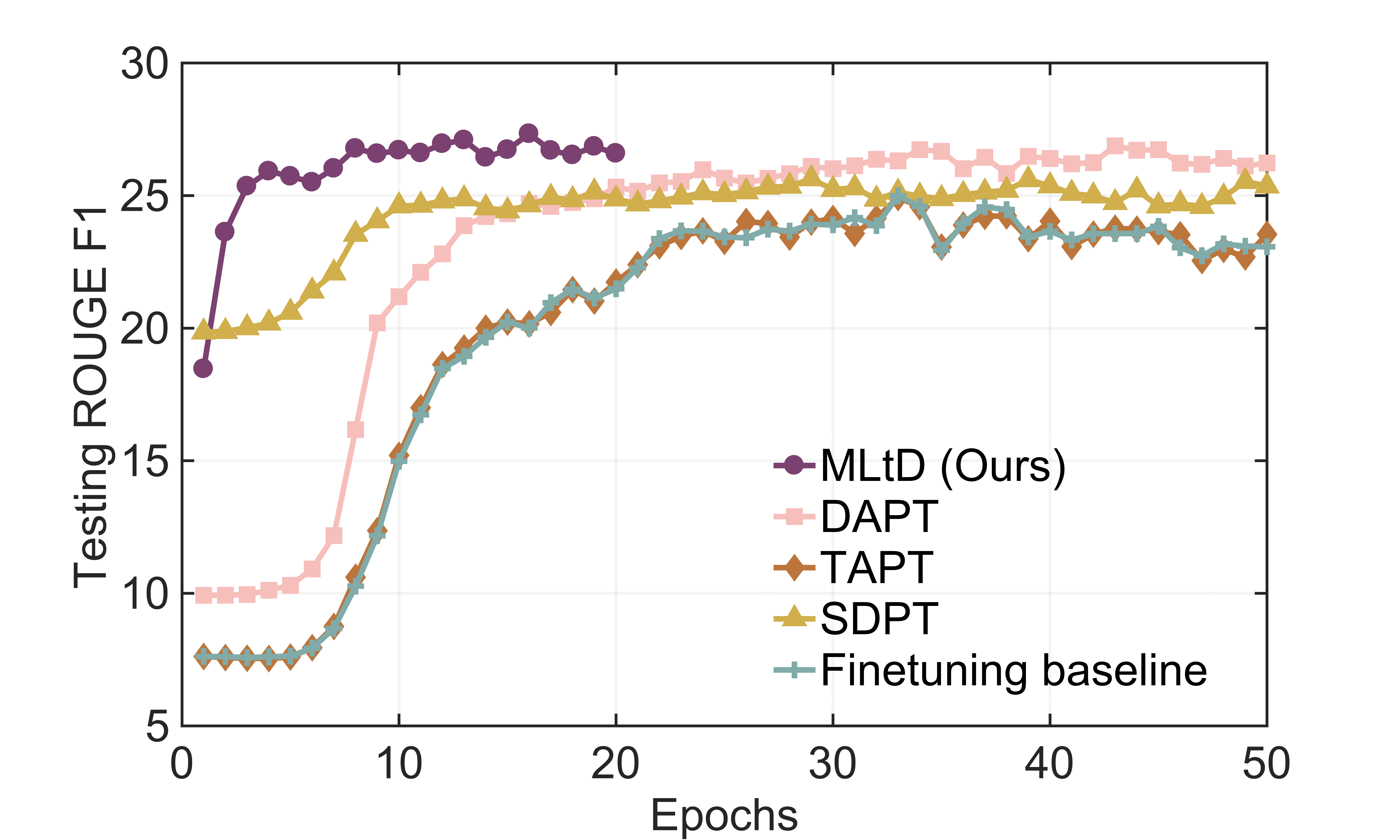}
\caption{Convergence analysis for finetuning BART models obtained by different methods on AdaptSum, using the Debate domain as an example.}
\label{fig:convergence}
\end{figure}

We have the following observations:
(1) since meta-learning explicitly optimizes the model for fast adaptation.  Compared with previous methods, MLtD takes fewer epochs to reach convergence (e.g., \Cref{fig:convergence}) and takes the least time to adapt the model to each the task; (2) since our lightweight adaptation method (TARP) only updates a small set of task-specific weights, our model variant (TARP only, no meta-learning) still reduces adaptation time by 25\% over direct BART finetuning.

On the other hand, the proposed TARP and TAMS components introduce some inference overhead.
Due to limitations of current DL libraries in implementing parallel computation branches, the dynamic low-rank decomposition and the task-aware architecture generation increases the inference time by 10\% and 6\%, respectively, measured with a batch size of 4 and a sequence length of 1024 on one GPU.

\section{Conclusion}
We have shown that explicit meta-learning is a useful preparation step on top of PLMs to improve later finetuning.
Specifically, our MLtD framework incorporating dynamic task-adaptive reparameterization (TARP) and task-adaptive model search (TAMS) enable data- and parameter-efficient adaptation to a family of low-resource tasks.
Future avenues include applying our method in other modalities like vision and speech, as well as exploring better model formulations for TARP and TAMS.

\section*{Acknowledgements}
We thank our colleagues on the Speech Science team at Amazon AWS AI for supporting this research.

{\small
\bibliography{anthology,custom}
\bibliographystyle{acl_natbib}
}

\appendix
\graphicspath{ {./figures/} }

\section{Further details for main experiments}
\label{sec:further-main}

\subsection{Hyperparameters}
\label{sec:hyperparams}

Most of the experimental setups, e.g., model type, max sequence, optimizer, batch size, beam search size, are taken from previous methods for fair comparison.
We tuned the inner-loop and outer-loop learning rates in meta-training on the meta-validation set, and adjust the learning rate schedule accordingly.
We chose the rank values $r$ in our dynamic low-rank reparameterization to give similar parameter counts to other parameter-efficient methods.
We adapted the search space in our task-aware model structure from DARTS.
$\eta_\text{in}$ denotes inner-loop and finetuning learning rate, $\eta_\text{out}$ denotes outer-loop learning rate, $B_\text{in}$ denotes inner-loop and finetuning batch size, $B_\text{out}$ denotes meta-batch size, $\text{bsz}$ denotes decoding beam size, and $T_\text{in}$ denotes inner loop steps.

\paragraph{Few-shot dialogue personalization.}
We take $r = 4$, $B_\text{out} = 16$ (as in previous works), $\text{bsz} = 5$, $T_\text{in} = 10$.
For meta-training we use SGD ($\eta_\text{in} = 0.01$) in the inner loop and Adam for the outer loop ($\eta_\text{in} = 0.0003$).

\paragraph{Low-resource abstractive summarization.}
We take $r = 16$, $B_\text{in} = 40$ (via gradient accumulation), $T_{\text{in}} = 20$, $\text{bsz} = 4$.
We truncated the input documents into 1024 tokens due to the limit of max input length of BART model.
We used Adam with momentum $(\beta_1=0.9, \beta_2=0.998)$ and the Noam schedule (linear warmup of 1000 steps, then inverse square-root decay).
Since the low-resource training set of science domain only has 100 samples, we used 3 times more training epochs than other domains. 

\paragraph{Multi-domain language modeling.}
We take $r=32$, $B_\text{in} = 4$, and $T_{\text{in}} = 1$. 
We used Adam with $\eta_\text{in} = 5 \times10^{-4}$, $\eta_\text{out} = 5 \times 10^{-5}$. In meta-testing we linear decay $\eta_\text{in}$.

\subsection{Training costs}
Table \ref{tab:amount_of_training} provides information about the amount of training that MLtD has taken for the main experiments. The table reports the training cost of a single run.
We tuned the hyperparameters, namely the inner-loop and outer-loop learning rates, over around five runs respectively.

\begingroup
\setlength{\tabcolsep}{4pt}
\begin{table}[h]
\centering
\scriptsize
\captionsetup{font=small}
\captionsetup[sub]{font=small}
\begin{tabular}{c c c c c c}\toprule
    \#GPUs & GPU & Training & Meta- & Cluster & Costs \\
    & type & time & iterations & & \\\midrule
    \multicolumn{6}{l}{\textit{Low-resource abstractive summarization (Table 1)}} \\
    1 & V100 & 39hrs & 100 & AWS p3.2xlarge & \$120  \\
    \midrule
    \multicolumn{6}{l}{\textit{Few-shot dialogue personalization (Table 2)}} \\
    1 & V100 & 1.5hrs & 100 & AWS p3.2xlarge & \$5  \\
    \midrule
    \multicolumn{6}{l}{\textit{Multi-domain language modeling (Table 4)}} \\
    1 & V100 & 22hrs & 100 & AWS p3.2xlarge & \$68  \\
    \bottomrule
\end{tabular}
\caption{Details on the training cost of MLtD. Cost (\$) estimated from on-demand instance prices.}
\label{tab:amount_of_training}
\end{table}
\endgroup

\section{Further details for TARP experiments}
\label{sec:adaptation-data}

\paragraph{Datasets.}

\textbf{E2E}
\citep{novikova-etal-2017-e2e} 
is commonly used for data-to-text evaluation of NLG systems. It consists of approximately 50K examples in total
from the restaurant domain. 
Each 
input consists of a sequence of slot-value pairs and can have multiple references. The average output length is 22.9. We use the official evaluation script, which reports BLEU, NIST, METEOR, ROUGE-L, and CIDEr.
\textbf{WebNLG} \citep{gardent-etal-2017-webnlg} is a multi-domain dataset for data-to-text evaluation.
It contains 22K examples in total from 14 distinct domains, and the average output length is 22.5. Nine domains are used for training, and the remaining five domains are used for testing. Each input is represented by a sequence of SUBJECT | PROPERTY | OBJECT triples. The evaluation metric is BLEU.
\textbf{DART} \citep{nan-etal-2021-dart} is an open-domain data-to-text dataset. The inputs are structured as sequences of ENTITY | RELATION | ENTITY triples. It contains 82K examples in total and the average output length is 21.6. The evaluation metric is BLEU.
\textbf{GLUE} We report Matthew’s correlation for CoLA, Pearson correlation for STSB, and accuracy for the other tasks in Table \ref{tab:sota_adaptation_glue}. 
The dev set performance is presented by following \citet{zhao-etal-2020-masking} and \citet{he2022towards}.

\paragraph{Setup.}
For the natural language generation tasks, we build upon \citet{hu21lora}'s code.\footnote{\url{https://github.com/microsoft/LoRA/tree/snapshot-9-15-2021}; they have greatly refactored their code since our experiments.} We used the GPT2-medium as the underlying LM. In training, we used the AdamW optimizer with weight decay 0.01. The batch size is set to be 8 and we trained for 5 epochs in total. We used linear decay learning rate scheduler with the first 500 iterations for warmup. The initial learning rate is set to be 0.0002. In decoding, we used beam search with beam size 10.

For GLUE tasks, we built upon \citet{he2022towards}'s code\footnote{\url{https://github.com/jxhe/unify-parameter-efficient-tuning}}. 
Our experiments were performed on RoBERTa$_{\text{base}}$ model. 
We limited maximum length of a sentence (pair) to be 512 after wordpiece tokenization. 
We used the Adam optimizer with batch size 32, and trained for 10 epochs on each task. 
The learning rate is a hyperparameter to tune for different tasks over
$\{1,2,3,4,5\} \times 10^{-4}$,
with a linear warmup for the first 6\% of steps followed by a linear decay to zero.

\end{document}